\pdfoutput=1

\documentclass[11pt]{article}

\usepackage[final]{ACL2023}

\usepackage{times}
\usepackage{latexsym}

\usepackage[T1]{fontenc}

\usepackage[utf8]{inputenc}

\usepackage{microtype}

\usepackage{inconsolata}
\usepackage{graphicx}
\usepackage{float}
\usepackage{titlesec}
\usepackage{listings}
\title{StyloMetrix: An Open-Source Multilingual Tool for Representing Stylometric Vectors}

\author{Inez Okulska \and Daria Stetsenko \and Anna Kołos \and Agnieszka Karlińska \and \\
\textbf{Kinga Głąbińska} \and \textbf{Adam Nowakowski}\\
NASK National Research Institute \\ Warsaw, Poland \\
\texttt{(inez.okulska, daria.stetsenko, anna.kolos, agnieszka.karlinska)@nask.pl}} 

\begin{document}
\maketitle
\begin{abstract}
This work aims to provide an overview on the open-source multilanguage tool called StyloMetrix. It offers stylometric text representations that cover various aspects of grammar, syntax and lexicon. StyloMetrix covers four languages: Polish as the primary language, English, Ukrainian and Russian. The normalized output of each feature can become a fruitful course for machine learning models and a valuable addition to the embeddings layer for any deep learning algorithm. We strive to provide a concise, but exhaustive overview on the application of the StyloMetrix vectors as well as explain the sets of the developed linguistic features. The experiments have shown promising results in supervised content classification with simple algorithms as Random Forest Classifier, Voting Classifier, Logistic Regression and others. The deep learning assessments have unveiled the usefulness of the StyloMetrix vectors at enhancing an embedding layer extracted from Transformer architectures. The StyloMetrix has proven itself to be a formidable source for the machine learning and deep learning algorithms to execute different classification tasks.
\end{abstract}

\section{Introduction}

Parallel to the rise of large language models and complex transformer architectures, there is a growing interest in widely explainable solutions, aligning with the trend towards responsible and transparent AI. Soon, model explainability will not only reflect the curiosity of researchers or an engineer's goodwill towards users, but will also become a top-down requirement, for instance, within the EU due to the forthcoming AI Act. Domain knowledge is simultaneously gaining prominence, as it is the key to a profound understanding of data and emerging patterns. Moreover, as a result, it enables an informed choice of data representations that support the desired transparency and facilitate explainability. 

StyloMetrix, the open-source multilanguage stylometric tool presented in this paper, combines both approaches – the potential for model explainability (the interpretability of the features), and the utilization of domain expertise. It allows for highly effective feature engineering and expert analysis of the results. \textbf{Interpretable StyloMetrix vectors serve two functions – 1) they can be used as input to explainable classification models and 2) they enable new knowledge discovery through stylometric analysis of the corpora included in the model's classes}. Previous experiments have demonstrated the effectiveness of these vectors in content (including malicious content) classification, genre identification, style analysis, and authorship attribution. Combined with neural embeddings like BERT-based models, stylometric vectors enhance classification accuracy. 

\textbf{Currently, StyloMetrix is available in four languages: English, Polish, Ukrainian, and Russian}, but its design allows for the rapid and convenient expansion of this set to include additional languages.

\section{Related Work}

Stylometry (or computational stylistics) is a broad field with a number of studies that involve the analysis of linguistic features extracted from a collection of texts in order to characterize the style of an author, document, or group of documents~\cite{eder2014}. The use of statistical techniques makes it possible to draw out the often subtle differences and similarities between texts that are invisible to the naked eye, and thus to delineate groups of texts based on their degree of linguistic affinity~\cite{piasecki2018}. Stylometric methods are successfully used in author authentication (e.g.~\citet{jankowska2014author},~\citeauthor{shrestha2017}), text classification (e.g.~\citeauthor{brocardo2013authorship}, ~\citeauthor{jockers2008reassessing}, ~\citeauthor{koppel2009computational}), active author verification (e.g.~\citeauthor{fridman2015multi}, ~\citeauthor{gray2011personality}), genre analysis (e.g.~\citeauthor{sarawgi2011gender}, ~\citeauthor{stamatatos2000automatic}), and other tasks. 

The development of methods and approaches in these fields has translated into the development of systems supporting stylometric analysis to a limited extent. Worth mentioning here are ‘Stylometry with R’ (stylo)~\cite{eder2016} and WebSty~\cite{piasecki2018, eder_open_2017, walkowiak_language_2018}, open tools that enable quantitative analysis of texts in various languages, including Polish. The stylo is a user-friendly and straightforward method for unsupervised multivariate analysis. The classification function of the tool is developed using supervised learning. WebSty is an easily accessible open-sourced tool for stylometric analysis and a part of the CLARIN-PL research infrastructure. WebSty covers a wide range of languages, provides grammatical, lexical, and thematic parameters to analyze the text, and enables the user to select relevant features manually. However, unlike the StyloMetrix vector, WebSty lacks the metrics for syntactic structures and usability. The tool is only available online through a Website with a limited range of pre-set options. 

Machine learning is used extensively in stylometric text analysis and authorship attribution. The word2vec algorithm ~\cite{mikolov2013efficient} looks at the surrounding context to construct word embeddings, while GloVe builds a more comprehensive vector based on the global statistics of a word co-occurrence~\cite{pennington2014glove}. FastText~\cite{bojanowski2017enriching} is another popular method where a word vector is constructed as the sum of associate character n-grams. These approaches of word embeddings are further extended to the document embeddings (Doc2Vec, ~\cite{le2014distributed}) and are commonly used in modern text analysis. However, their interpretability remains unclear and seems like a black box to the user. We share Yu’s~\cite{bei2008} perspective on the feature extraction process. The scholar claims that the feature vector should be plausible and easily interpretable before it is employed in any machine learning or deep learning algorithm.

Therefore, the purpose of the StyloMetrix is primarily to be a syntactic and lexical vector representation that can be interpreted by any linguist expert. The developed metrics are common and widely utilized in the linguistic community, however some of them are more elaborate and task-specific, which enables a user to choose the exact set of metrics that is needed for their purposes.

Although the StyloMetrix is one of the rare multilingual comprehensive corpus analysis tools, the first commonly used apparatus for an exhaustive readability analysis was developed in 2006. The tool called Coh-Metrix\footnote{\url{http://cohmetrix.memphis.edu/cohmetrixhome/}} was developed by a group of four researchers from the Institute for Intelligent Systems, The University of Memphis, and released to the public via an online platform. In September 2022, the Coh-Metrix project was revived and transformed into Coh-Metrix 3.0 \cite{mcnamara_graesser_mccarthy_cai_2014}. The number of metrics has increased to 200. However, the size of an analyzed text is still limited to 1000 words. Moreover, Coh-Metrix offers a choice of statistical estimation (mean or standard deviation). Coh-Metrix encompasses a wide array of metrics covering grammar, syntax, and semantics of the English language. It analyzes texts on over 50 types of cohesion relations and over 200 measures of language, text, and readability. CohMetrix has been widely used by about 6000 researchers from all over the world. It has also been a source of inspiration for analyzing languages other than English: there are documented versions of CohMetrix for Spanish \cite{quispesaravia-etal-2016-coh}, Portuguese\footnote{\url{fw.nilc.icmc.usp.br:23380/cohmetrixport}}, and Chinese \cite{Ye2013ACA}, however no versions tailored for Ukrainian, Polish, or Russian languages have yet been developed.
The StyloMetrix metrics described further in this paper share a conceptual similarity but diverge in their intended applications. While Coh-Metrix primarily focuses on coherence, cohesion, and readability scores, StyloMetrix vectors are designed to encompass a wide and unconstrained spectrum of grammatical and syntactic distinctions found across different text genres. 

\section{StyloMetrix: Architecture and Design}
Drawing inspiration from stylistic features for stylometric analysis, we have developed metrics grounded in grammar, syntax, and lexis. Each metric calculates the total count of tokens adhering to specific linguistic rules. The outcome of the StyloMetrix tool is a normalized vector for each input text. This feature facilitates the comparison of texts of varying lengths within the same genre or those authored by the same individual, among other potential use cases. The StyloMetrix Python package is readily available as an open-source tool on GitHub\footnote{\url{https://github.com/ZILiAT-NASK}}, accessible to all. Furthermore, the tool's architecture is designed to be customizable by computational linguists, who can either modify existing metric sets or introduce their own.

\begin{figure}[H]
\includegraphics[width=\columnwidth]{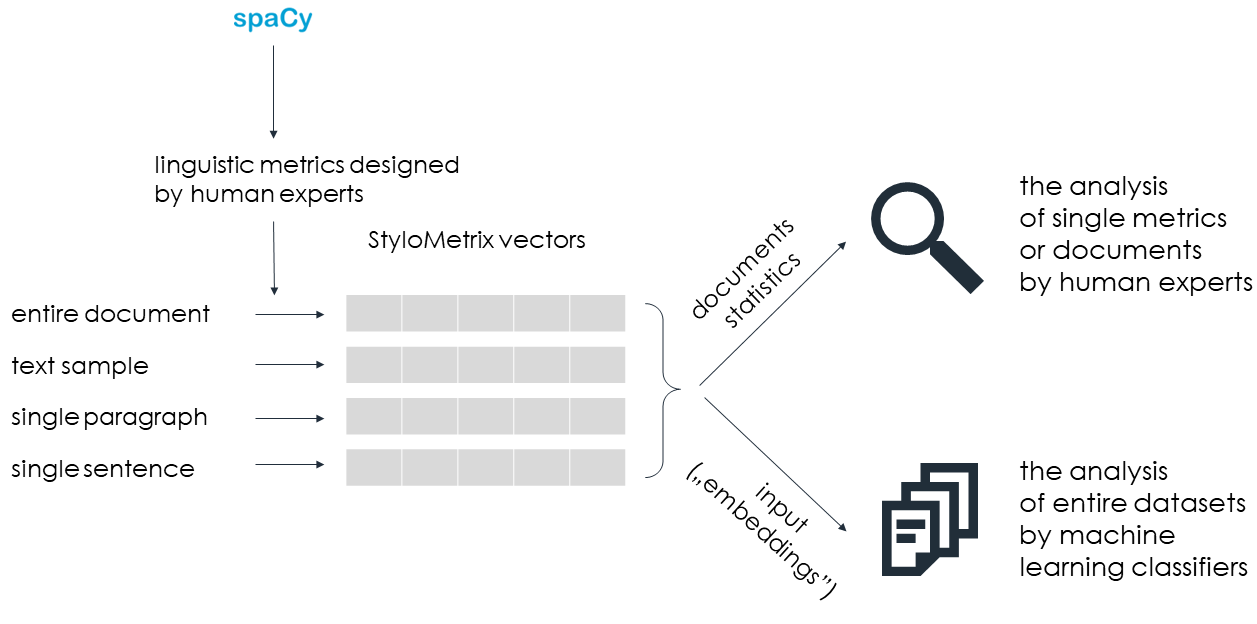}
\caption{The concept of StyloMetrix}
\label{fig:sm}
\end{figure}

The foundation of the StyloMetrix tool is rooted in an open-source spaCy model developed by Explosion AI~\cite{spacy2}. SpaCy models have established themselves as front-runners, showcasing state-of-the-art performance across a spectrum of tasks including tokenization, parsing, tagging, and named entity recognition, among others. The English, Ukrainian, and Russian StyloMetrix pipelines draw upon the output generated by spaCy models. Unlike those, the Polish StyloMetrix utilizes a model developed by the Institute of Computer Science (Polish Academy of Sciences), which incorporates a wider range of morphological features, compared to the standard spaCy version. Further, the attributes such as part of speech label, tag and a set of morphological characteristics are leveraged to create higher level syntactic and grammatical metrics. The input text is analyzed according to the hand-crafted rules and settings. As a result, as an output the user receives a csv file with normalized statistics for each metric. See: fig.\ref{fig:sm}. 

\section{Potential for Explainability}

A pivotal objective ingrained in the design of StyloMetrix is the interpretability within the vectors it generates. Interpretation is possible at different levels, depending on the specific task or the level of technical expertise of the users. Beyond delivering normalized statistics, the tool extends its utility by providing debug files as an auxiliary output format. These debug csv files encapsulate each token captured by each metric, allowing the user to delve deeper into linguistic analysis. 

\begin{figure}[!ht]
    \centering
    \includegraphics[width=\columnwidth]{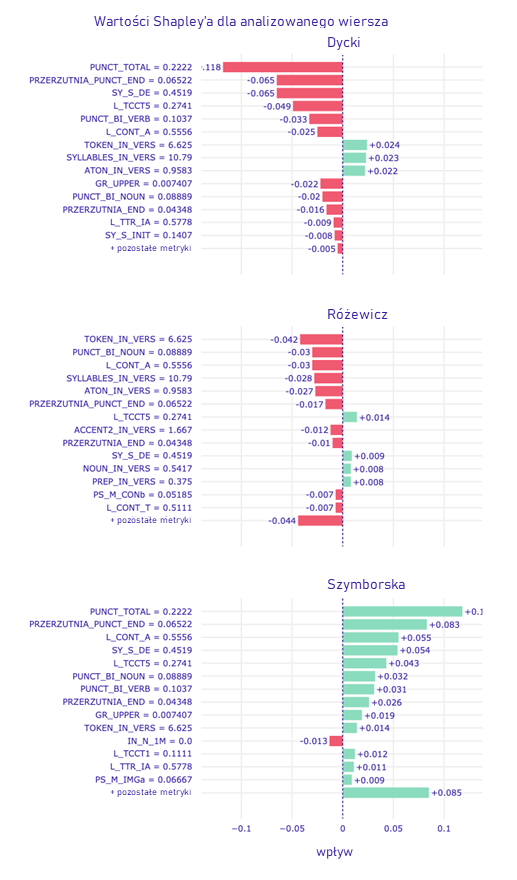}
    \caption{Shapley values for the Stylometrix features visualized with the Dalex package in a classification task of Polish poetry. StyloMetrix works effectively with easy-to-explain models such as Random Forest. And since the features translate directly into specific grammar-related patterns, explaining model's decision allows also for linguistic interpretation of the analyzed samples.}
    \label{fig:xai}
\end{figure}

For comparative analysis in computational linguistics, statistical tests can be used to determine the significance of different features in discriminating between certain groups of texts. For the different tasks of text classification, both the most important features and the correlations between pairs of these features can be used to gain insight into the model's decision making. To explain the model's decisions in a more complex way, i.e. taking into account the interaction of pairs of features and also the decomposition of the model's decisions into features that influenced the decision in a certain way or introduced uncertainty, one can use the open source Artemis library\footnote{\url{https://github.com/pyartemis/artemis}}. It builds upon the Dalex package \cite{JMLR:v19:18-416} that is also suitable for visualizing other type of model explanation such as Shapley Values (see fig. \ref{fig:xai}). 

\section{StyloMetrix Metrics}
\subsection{Metrics for Polish}

The Polish language, which is classified as a Western or sometimes Northern Slavic branch of the wider Indo-European language family, is distinguished by its synthetic and inflectional or fusional nature \cite{smoczynska_acquisition_1985, reid_organising_2000, LewandowskaTomaszczykWilson}. A comprehensive set of 172 metrics has been developed specifically for the Polish language. These metrics encompass various linguistic aspects, including: \begin{enumerate}
  \item Grammatical forms
  \item Punctuation
  \item Syntax
  \item Inflection
  \item Graphical representation
  \item Lexical attributes
  \item Psycholinguistic features
  \item Descriptive characteristics
\end{enumerate}

All of these metrics have been meticulously crafted by experts to address the unique challenges faced by morphologizers, parsers, and taggers based on the spaCy framework. The linguistic information associated with each token is structured across six levels: (1) Lemmas (the base or root forms of words), (2) Part-of-Speech (POS) tagging (assigning grammatical categories to words), (3) Morphology (examining the internal structure and form of words), (4) Tags (utilizing the National Corpus of Polish tagset for tagging)\footnote{The National Corpus of Polish (NKJP): \url{http://nkjp.pl/}. See: \citet{nkjp2012}.}, (5) Dependency (analyzing the syntactic relationships between words), (6) Entity label (identifying named entities and categorizing them).
The choice of which linguistic description levels to consider in the programming code depends on the specific metric type. This strategic approach aims to capture the intended linguistic properties effectively while mitigating potential inconsistencies arising from errors made by the spaCy tools. 

\subsubsection{Grammatical Forms}

The Grammatical Forms class has been crafted to encompass the prevalence of distinct parts of speech. This comprehensive classification begins with nouns, verbs, adjectives, adverbs, and extends to a spectrum of diverse pronouns, including personal, demonstrative, possessive, total, interrogative, indefinite, and negative ones. Additionally, it encompasses particles, adpositions, coordinating and subordinating conjunctions, numerals, collective numerals, interjections, symbols, abbreviations, and other linguistic elements, primarily targeting foreign lexemes that may not be readily recognized by Polish language processing tools.

\subsubsection{Inflection}

The most extensive class is primarily dedicated to the intricate aspects of inflection, which are of utmost importance to the Polish language. This category encompasses a total of 64 distinct metrics, covering the nuances of noun and pronoun declension, the comparison of adjectives and adverbs, and the conjugation patterns of verbs, providing a comprehensive framework for the analysis of grammatical inflection. 7 metrics are focused on the number and gender of nouns, encompassing both masculine and non-masculine personal genders. In addition to addressing the standard verb conjugation, this set also encompasses 24 metrics devoted to specific properties of verbs. These  encompass a wide range, including the presence of finite and infinitive verbs, the utilization of present, past, and future tenses, the application of active and passive voices, the consideration of perfect and imperfect aspects, and the incorporation of imperative and conditional moods. Additionally, noun-like gerunds (nominalized verbs), quasi-verbs (mostly impersonal non-inflected forms functioning like verbs), distinct types of participles (adverbial and adjectival ones) and impersonal verb forms in perfective and imperfective aspect are included. 

\subsubsection{Syntax}

In the present iteration of StyloMetrix, the Syntax class encompasses a total of 19 metrics. However, it is essential to note that the development of syntax-oriented metrics for Polish is an ongoing endeavor. This ongoing effort is driven by the inherent challenges posed by the complex and flexible nature of Polish syntax, which lacks strict rules governing word order. Additionally, the intricacies of dependency tagging for Polish further contribute to the complexity of this task. In the current state, these metrics are designed to analyze various aspects of longer text units. They include the identification of words within declamatory, exclamatory, interrogative, negative, ellipsis-ended, and nominal sentences, as well as words enclosed in quotation marks. Specific considerations are made for colloquial speech, particularly in instances where infinitive verbs are used to convey an imperative-like intention. Acknowledging the unique characteristics of Polish syntax, the metrics encompass nominal predicates, Object-Verb-Subject (OVS) word order, and inverted epithets. Additionally, they extend to cover nominal phrases, words situated within modifiers, flat multiword expressions (FME), and appositional modifiers. Two additional metrics have been devised to identify similes, with one specifically targeting constructions in which nouns and pronouns function as objects of the comparison. The other metric is dedicated to the analysis of adjectival comparisons. An additional set of metrics, specifically concentrating on subordinate and compound sentences, including various types of adverbial clauses such as causal, conditional, and concessive clauses, is currently in development and being closely monitored. Once the testing phase is successfully completed, these metrics will be made available on our GitHub account for public access and use.

\subsubsection{Lexical attributes}

A total of 34 metrics have been designed to delve into the lexical aspects of texts. Among these, 12 metrics cover proper names, including masculine and feminine forms, and named entities identified using the Named Entity Recognition (NER) component integrated with the Polish spaCy model\footnote{\url{https://spacy.io/api/entityrecognizer}}. These metrics cover a wide range of entities, including person names in masculine and feminine forms, organization names, and dates. Furthermore, these metrics exhibit a nuanced approach by combining entity labels with animacy status, as provided by morphological analysis. This combination enables the creation of distinct metrics that can capture place and geographical names on one hand, and ethnonyms and demonyms, which are lexemes related to humans, on the other hand. Additionally, the metrics encompass a category of adjectives derived from place and geographical names.

15 lexical metrics have been curated to capture the diversity of the vocabulary used in texts. Among these metrics, four are dedicated to the measurement of word length, a characteristic quantified by syllable count. For this purpose, we used Spacy Syllables\footnote{\url{https://spacy.io/universe/project/spacy_syllables}}, another component of the SpaCy pipeline that adds multilingual syllable annotations to tokens\footnote{It is worth noting that word length, measured in syllables, is considered an important indicator of text comprehensibility and is used in various readability measures (usually the ratio of words with 3 or more syllables to all words or, in the case of measures developed specifically for Polish, the ratio of words with 4 or more syllables to all words is often the basis for estimating the level of text difficulty, see \citet{Broda2014, Gruszczyński2015}. In future iterations of StyloMetrix, various readability measures will be incorporated within the lexical metrics class}. Concurrently, other lexical metrics are devoted to the distinction between content words and function words, considering both lemmatized and non-lemmatized forms. Furthermore, a dedicated metric scrutinizes the prevalence of stop words, drawing from the spaCy stop words list tailored for the Polish language. The Type-Token Ratio, crucial for assessing the lexical richness and diversity of a text \cite{Richards1987}, reflected in the ratio between tokens and word classes (types) in a corpus, is analyzed in two dimensions, including both lemmatised and non-lemmatised tokens. Additionally, two metrics facilitate the identification and quantification of the 1\% and 5\% most frequently occurring token types within the text.

Within this subset of 6 lexical metrics, a dictionary-based approach is employed for linguistic analysis. The metric for vulgarisms focuses on the detection of vulgar language and expressions by utilizing a predefined dictionary containing inflected words derived from the most common formative Polish vulgar roots. Another metric is designed to pinpoint frequently occurring linguistic errors and relies on a dedicated dictionary that has been compiled in reference to an annual report detailing the "100 most frequent errors on the Internet."\footnotetext{\url{https://nadwyraz.com/blog/raport-100-najczesciej-popelnianych-bledow/-w-internecie-w-2021}. The annual report is curated by web services www.nadwyraz.com and www.polszczyzna.pl in partnership with SentiOne.} The growing prevalence of Greek-origin prefixes, such as \textit{hiper}, \textit{mega}, or \textit{super}, which primarily intensify adjectives but can also affect other parts of speech, is discernible through the lexical metrics. To accommodate the diversity of non-standard word formation spellings, apart from defining a comprehensive list of these intensifiers, an auxiliary dictionary of exceptions was compiled to exclude fossilized lexemes that commence with prefixes like \textit{giga}, etc.\footnotetext{The auxiliary dictionary was sourced from Great Dictionary of Polish (WSJP). See: \url{https://wsjp.pl/}} Tracking the occurrence of fixed adverbial phrases within the text is also possible owing to a list of such phrases sourced from Wiki słownik, a component of Wiktionary\footnotetext{\url{https://pl.wiktionary.org}}. Furthermore, three dictionary-based metrics account for the incidence of adverbs of time, duration, and frequency.

\subsubsection{Psycholinguistic features}

Psycholinguistic features related to emotional affect were extracted from an experimental study on affective norms for Polish words ~\cite{imbir2016, imbir2023}. A dictionary comprising 2,650 words, developed by Imbir et al., was employed. In the study, individual words underwent evaluation by a total of 1,380 participants using self-report scales in three two-dimensional spaces: valence (which encompassed dimensions of positivity and negativity), origin (including automaticity and reflectiveness), and activation (involving subjective significance and arousal). In designing the metrics for the Psycholinguistic class, we took into account the six dimensions emphasised in the study, with each dimension further broken down into two metrics: the first metric counted the proportion of words exceeding the mean value for a given feature (e.g., positivity), while the second metric counted the proportion of words falling below the mean value for a given feature. In total, 12 psycholinguistic metrics were created.

\subsubsection{Descriptive characteristics}
The Descriptive Characteristics class was designed to detect intricate linguistic patterns, encompassing the precise utilization of adjectives and adverbs in the descriptions of qualities, the examination of adverb pairs and phrases where an adverb precedes an adjective. This category also places particular emphasis on complex apostrophes, incorporating either adjectives or verbs, as well as encompassing longer nominal phrases. Furthermore, it accounts for the distinctive use of nouns and pronouns in the vocative case.

\subsubsection{Punctuation and the Graphical representation}

The Punctuation and Graphical Representation class addresses various aspects of the textual structure. The punctuation metrics are tailored to record the frequency of punctuation marks, also in proximity to nouns or verbs. On the other hand, the graphical properties metrics focus on identifying capitalized tokens and recognizing distinctive features commonly found in social media texts, such as emojis, emoticons, lenny faces, URLs, hashtags, and mentions with "@" symbols. For the detection of emojis comprising one or more Unicode characters, we employed the spacymoji, an extension and pipeline component integrated with spaCy \footnote{\url{https://pypi.org/project/spacymoji}}. Emoticons were identified using the Emoticon Dictionary, comprising 225 of the most frequently used emoticons along with their textual representations \cite{baziotis-pelekis-doulkeridis:2017:SemEval2}. Lenny faces, URLs, hashtags, and mentions with "@" symbols were extracted through the application of custom-designed regular expressions.

The succinct overview of distinct metric groups already underscores the notion that various features have been tailored to capture characteristics specific to different types of texts and genres. While metrics aimed at grammatical forms, inflection, or syntax are universally applicable to all types of texts, spanning from literary to practical, others have been devised to detect attributes unique to social media posts (as exemplified by those encompassed in Graphical Representation) or, in a broader context, to informal speech patterns, encompassing online discourse. While the presented set of metrics is by no means exhaustive, particularly in its coverage of a plethora of non-normative linguistic patterns, a substantial number of metrics have been thoughtfully curated by linguistic experts who have drawn upon their extensive experience in the analysis of informal Polish discourse, encompassing both commercial and non-commercial blog posts, as well as offensive or harmful content on social media platforms.

\subsection{Metrics for English}
The StyloMetrix model for English is built upon the SpaCy~\cite{spacy2} English transformer pipeline (RoBERTa-based) with components: transformer, tagger, parser, NER, attribute ruler, and lemmatizer. The choice of the model is driven by the accuracy evaluation. The English version of the StyloMetrix leverages POS Tags by the PennTree Bank, dependency labels, and morphological features. Although the SpaCy English parser performs well, we encountered some drawbacks while designing the metrics. To implement the extension for the present tenses, we take into consideration the incorrect sentence parsing. For instance, in the sentence: \textit{Diego's trying to splash water onto her back}, -- "s" is tagged as the auxiliary verb in the passive voice; when it should be the auxiliary verb in the active voice. The SpaCy English transformer pipeline performs well on sentences without contractions, where each word is disambiguated. Moreover, while the sentence length increases, the parser's performance decreases due to the long-distance dependencies and projectivity. 

We found it essential to compare existing parser tools with the spaCy transformer, to assure that spaCy provides the most rigorous results. To perform the comparative analysis we applied Stanford~\cite{de2006generating}, Berkeley~\cite{johnson2010reranking} and SpaCy English parser. As predicted, the spaCy parser showed the best performance on the task as a word disambiguation, where the sentence incorporates the contractions such as \textit{we'd} or \textit{I'd}, that can be parsed in two ways: as a verb \textit{had} or a modal verb \textit{would}.

After the preliminary stage of parser and tagger analysis we have crafted the metrics based on the same principles which were applied for the Polish language version. Hence, we will not repeat the shortcomings and obstacles acquainted during the process of creating the rules. In total the English version covers 196 metrics, divided by groups:
\begin{enumerate}
  \item Detailed grammatical forms
  \item General grammar forms
  \item Detailed lexical forms
  \item Additional lexical items
  \item Parts of speech
  \item Social media
  \item Syntactic forms
  \item General text statistics
\end{enumerate}

\subsubsection{Detailed and general grammatical forms}
The detailed grammar group is also the most extensive one. It incorporates 55 rules which cover most of the English tenses. That is, it encompasses the present, past and future tenses in their various verb forms and two aspects: the present / past / future simple; continuous; perfect; perfect continuous active or passive aspect. The modal verbs also belong to this groups. The StyloMetrix covers most of the frequently used modal verbs, such as: can / could, may / might, shall / should, must and would. Each of them are evaluated in the present, continuous, perfect forms; active and passive voice. 

Pervasively, we rely on the spaCy morphology and dependency parsers and implement extensions for each rule, which allow to add metrics as the custom component to the main pipeline.

General grammar forms are the consolidation of the principal grammatical rules. Under this category falls the present tenses; the past tenses; the future tenses; infinitive forms; modal verbs in the simple form; modal verbs in the continuous form; modal verbs in the perfect form; verbs in the active voice and verbs in the passive voice. Therefore, a user is able to choose which group of grammar metrics is more pertinent to their task. 

\subsubsection{Detailed lexical forms and Additional lexical items}
The second largest group is the detailed lexical. There are 48 metrics that mostly cover different types of pronouns. The subsets include subject, object, possessive, reflexive pronouns and four groups that incorporate the first person singular pronouns, the second person pronouns, the third person singular and plural pronouns. During the experiment stages we found that some lexical items contribute to the decisions made by a classification model to discern text genres, detect hate speech and abusive language, hence the hurtlex dictionary was added to the group of additional lexical items\footnote{\url{https://github.com/valeriobasile/hurtlex}} which encompasses a range of rude, abusive and hurtful words that are present on the mass media platforms. These groups of lexemes help to differentiate abusive or offensive language from non offensive. Indeed, this is the separate subgroup and can also be turned on or off based on the user's desire.
To the additional lexical group belong punctuation (the same instances as described for the Polish language), retweets, urls, mentions with an "@" sign, hashtags, content words and function words. Plural, singular, proper nouns, personal names, noun phrases and three forms of adjectives and adverbs constitute this group as well. Furthermore, we created the dictionary which helps to calculate words that denote common patterns such as: time and location; manner; cause and purpose; condition; limitation and contradiction; example; agreement and similarity; effect and consequences. We have observed that these linking expressions are useful for genre and author identification and can be a worthwhile extension to the existing lexical sample.

\subsubsection{Parts of speech and Social media}
Parts of speech and Social media are not as numerous as previous examples. The first is primarily oriented to calculate the general frequency of a specific POS in the text. These 23 metrics fully rely on the POS tagger by spaCy which, in turn, utilizes the Penn Treebank universally standardized annotation.
The social media group consists of 7 metrics, where two of them calculate positve or negative sentiment of the text by  an open-sourse library VaderSentiment\footnote{\url{https://vadersentiment.readthedocs.io/en/latest/}}. Other metrics aim to trace the lexical intensifiers, e.g. \textit{utterly}, \textit{tremendous}, and nomenclature words, e.g. \textit{occasionally}, \textit{little}, \textit{marginally}. examples of masked words (e.g. f**k) and digits (e.g. 4 u) also belong to the social media part.

\subsubsection{Syntactic forms and General text statistics}
22 metrics comprise the syntactic group which covers five types of questions: general, special, tag, alternative and negative; coordinate and subordinate sentences and number of punctuation marks in them; narrative and negative sentences as well as a direct speech. Some figures of speech are added to the group. Fronting refers to a construction where a group of constituents that usually follows a verb preceeds it instead. This figure of speech is also called preposing or front-focused. For instance, \textit{"Carefullt, a baby is sleeping."} or \textit{"On the corner stood a little shop."}\footnote{\url{https://dictionary.cambridge.org/grammar/british-grammar/fronting}}. Syntactic irritation is built upon the rule, that continuous form of any tense together with intensifiers such as "constantly", "continuously", "always", "all the time", "every time" -- create the effect of irritation or dissatisfaction. For example, \textit{"She's always coming late."}; \textit{"He was constantly losing his temper in public"}. Syntactic intensifiers affirm the sentence meaning and are presented in the form of auxiliary verbs: do, did, and does. An example of it we can find in the sentence \textit{"I do love dogs"}. Simile as a figure of speech is also implemented in this group, it covers instances such as: \textit{She looks like her mother.}; \textit{"He's as busy as a bee"}. Inverse sentences aim to catch such statements as \textit{"You will find only what you bring in."}; \textit{"Once you start down the dark path, forever will it dominate your destiny"}\footnote{\url{https://parade.com/943548/parade/yoda-quotes/}}.

General text statistics incorporates 12 metrics. A large number of them are designed to calculate the distance between the specific nodes. For instance, the distance between noun, verb, adverbial, prepositional and adjectival phrases within one text. Three types of the type-token ratio intend to show the readability score of the text and its cohesiveness. Repetitions of words and sentences also serve as markers of a text cohesion.

\subsection{Metrics for Ukrainian and Russian}
The sets of metrics for the Ukrainian and Russian languages are at the development stage, where at this moment they encompass 104 metrics
These metrics are subdivided into four main clusters:
\begin{enumerate}
  \item Lexical forms
  \item Parts of speech
  \item Syntactic forms
  \item Verb forms
\end{enumerate}
There is one additional group for the Ukrainian StyloMetrix that covers readability scores which are identical to the English version, so we will not concentrate much attention on them. 

As Ukrainian and Russian are fusional languages they have similar syntactic structures and grammatical characteristics. The main goal of developing both languages is to conduct the analysis on disinformation detection, as it is one of the most relevant tasks in NLP these days. 

The set of metrics for the Ukrainian and Russial languages are built based on the same principles and assumptions that were applied for the Polish and English languages. Hence, the lexical metric provide the information of plural and singular nouns, moreover we extended them to cover more morphological features such as animacy (animate/inanimate), gender(feminine, masculine and neutral), distinguish between first and second name. We also added diminutives, as they serve as a distinctive feature of these languages. Direct and indirect objects as well as cardinal and ordinal numerals are in the general set. We strive to pinpoint the distinctive lexical forms such as seven cases in Ukrainian vs six in Russian; demonstrative, personal, total, relative and indexical pronouns; as well as qualitative, quantitative, relative, direct and indirect adjectives. These are the unique features that bring not only semantic but also syntactical information about the text and help to classify genre peculiarities.
Other lexical subgroups as punctuation, direct speech and three types of adverb and adjective comparison are similar to the English version.

The Parts of speech class relies on the assumptions utilized by Polish and English languages. 

On the syntactic level we created an extension for the noun phrases, a common rule to detect the direct speech as well as narrative, negative, and interrogative sentences. 

One of the most prominent figures of speech that can be traced in both languages are parataxis, ellipsis and positioning. Parataxis is a syntactic mean that is based on the omission of conjunctions\footnote{\url{https://www.litcharts.com/literary-devices-and-terms/parataxis}}. For example, \textit{"I came, I saw, I conquered."} instead of \textit{"I came, and I saw, and I conquered."}. Ellipsis is an omission of words, represented as three dots, it aims to show a pause, or suggests there’s something left unsaid\footnote{\url{https://www.grammarly.com/blog/ellipsis/}}. For instance,\textit{ "She remained silent ... then things suddenly changed."}. Typically, ellipses are common in poetic or fictional texts and, hence, are a fruitful way of a genre discretion. Positioning is pertinent only to the Ukrainian language, it's a special way of constructing a phrase, where the first part (usually an adjective) describes the second one (usually noun). Unlike other languages, in Ukrainian such cases are written using a dash (-) and count as obsolete units. The described figures of speech are efficient at the classification tasks for author identification or genre differentiation. An example will be provided in the next section.

The last group consists mainly of the verb forms which cover past, present and future tenses forms as well as verbs in perfect and imperfect aspects, transitive and intransitive forms, and participles. The Ukrainian language varies from Russian due to the presence of four conjugation groups which are constructed as custom extensions in the pipeline. Moreover, incidence of adverbial perfect and imperfect participles are built for the Ukrainian StyloMetrix. 

As it was mentioned earlier, the Ukrainian and Russian versions of the StyloMetrix are in their preliminary stage and still developing. The main emphasis lies on the syntactic features, such as figures of speech, which are common and prominent for these languages. 


\section{Applications and Use Cases}

We designed the StyloMetrix to be a versatile text representation that can be utilized together with machine learning models. \textbf{The StyloMetrix output is calculated by the general statistical formula to estimate a mean value, i.e. the set of words which fall under a specific rule is divided by the total number of words in the text. Hence, each metric is always estimated in the range [0, 1]}.

The algorithm of obtaining the StyloMetrix logits is rather straightforward. One can download and install the tool via GitHub, then run the package on the local machine and generate the csv output. The inputs can be texts of any length. The StyloMetrix does nor require any additional preprocessing, e.g. url omission or hashtags elimination, as these features are covered on the metrics level. An important thing to remember is that the analyzed documents should be stored together in a separate folder for the tool to give a desired output. 

After generating the csv file a user can proceed to utilizing the StyloMetrix vector as inputs to the machine learning and deep learning algorithms.

In this section we present a couple of example settings that cover all languages provided by the StyloMetrix.

\subsection{Integration with Machine Learning Classifiers}
The StyloMetrics vectors prove themselves being robust and exhaustive mean to classify news genres and sources. These experiments we led by relying on the Ukrainian and Russian sources.

As Ukrainian is a low-resource language, and more datasets are yet to be developed; we have chosen a benchmark news corpus provided by ~\citet{panchenko2022ukrainian}. It was compiled of seven Ukrainian news websites:
BBC News Ukraine, New Voice Ukraine, Ukrainian
Pravda, Economic Pravda, European Pravda, Life Pravda, and Unian.

\begin{figure}[!ht]
    \centering
    \includegraphics[width=\columnwidth]{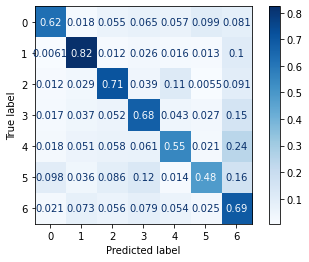}
    \caption{A confusion matrix for the Ukrainian news classification task}
    \label{fig:ukr}
\end{figure}

We strive to show how the StyloMetrix vectors can handle multi-class data by being executed only as an input to the Random Forest Classifier. Figure ~\ref{fig:ukr} displays a confusion matrix for seven news classes. We can deduce that most of the resources path the 50\% prediction threshold, which is a significant accomplishment of the StyloMetrix tool, paying attention that we have utilized a simple Random Forest classifier. It is worth noticing that as we dwell deeper into the texts that attain to every news resource, we have discovered that such classes as Ukrainian
Pravda and Economic Pravda belong to one publisher, therefore the style of writing is rather similar. Hence, we can see the confusion between the classes 4 (Ukrainian Pravda) and 6 (Economic Pravda).

To compare the output for the Ukrainian test we led a similar experiment with the Russian language. The benchmark news  corpus provided by the Huggingface \footnote{\url{https://huggingface.co/datasets/mlsum}} offers five classes that cover different news topics: culture, economics, politics, social, and sport. Unlike the Ukrainian experiment, this one focuses on the topic classification. The test settings remained the same: Random Forest classifier with the StyloMetrix vectors as an input. 

Taking into consideration that both languages capture similar features and syntactic patters we expect for the Russian model to perform on the same level as the Ukrainian model, regardless the difference in a task characteristics. 

\begin{figure}[!ht]
    \centering
    \includegraphics[width=\columnwidth]{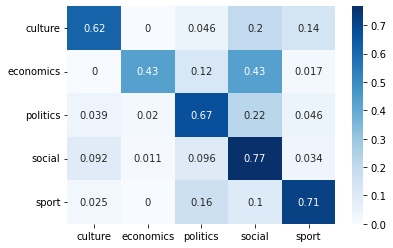}
    \caption{A confusion matrix for the Ukrainian news classification task}
    \label{fig:ru}
\end{figure}

Figure \ref{fig:ru} presents the outcome for the Russian news classification. The general tendency of the models performance is above the 50\% threshold except the economics class that gets confused with the social data. A high number of false positives can be driven by the similarity in syntactic constructions which are typically utilized in these types of texts. 

Overall, the StyloMetrix vectors prove their relevance for classification tasks in Ukrainian and Russian languages. Although, the total number of metrics for each language is quite low, only 104, they indeed help to differentiate the stylometric nuances of text genres and discern the source of each text.

The third experiment concentrated on the hate speech identification in the English language. A lot of experiments have been offered to differentiate offensive and neutral language, however most of them applied pervasively large Transformer or deep learning models. Unlike those, we tested the Voting Classifier on the Gab \cite{qian2019benchmark} dataset. The model is grounded on the Random Forest classifier, Logistic Regression and SVM algorithms. The StyloMetrix vectors managed to rich the overall performance close to the state-of-the-art deep learning models. The Table ~\ref{tab:vot} compares the Voting Classifier to the results of the deep learning algorithms obtained by \citet{qian2019benchmark}. As can be inferred from our experiment, the potential of the syntactic representations to distinguish hate speech from neutral is not a far-fetched idea, and can be executed with application of the simple rule-based approach like the StyloMetrix.

\newblock
\begin{table}[!ht]
\centering
\scalebox{0.8}{%
\begin{tabular}{|l|l|l|l|l|}
\hline
\textbf{Model} & \textbf{Prec.} & \textbf{Rec.} & F1 weighted & F1 hate \\ \hline
Voting: StyloMetrix & 0.83              & 0.84                 & 0.84        & 0.84    \\ \hline
CNN                                                        & 0.95              & 0.96                 & 0.90        & -       \\ \hline
RNN                                                        & 0.95              & 0.95                 & 0.89        & -       \\ \hline
\end{tabular}%
}
\caption{Voting Classifier with the StyloMetrix vectors for hate speech detection task.}
\label{tab:vot}
\end{table}

Speakleash library\footnote{\url{https://github.com/speakleash/speakleash}} is an open dataset for the Polish language systematically collected by the association Speakleash a.k.a Spichlerz, comprises over 300 GB of data (more than 54 million documents) from various categories. One of the significant advantages of this dataset lies in its immense diversity. Six categories were randomly selected (Job offers, Literature, News, Style blogs, Web articles and Wikipedia), with 30 texts randomly sampled from each. The Random Forest model was trained with 30 important features selected with the topK algorithm. 

Even with such a small number of samples, using StyloMetrix vector representation, these categories could be clearly distinguished with high accuracy, confirming the hypothesis that specific text types differ in grammatical structures and combinations.

\begin{figure}[h]
    \centering
    \includegraphics[width=\columnwidth]{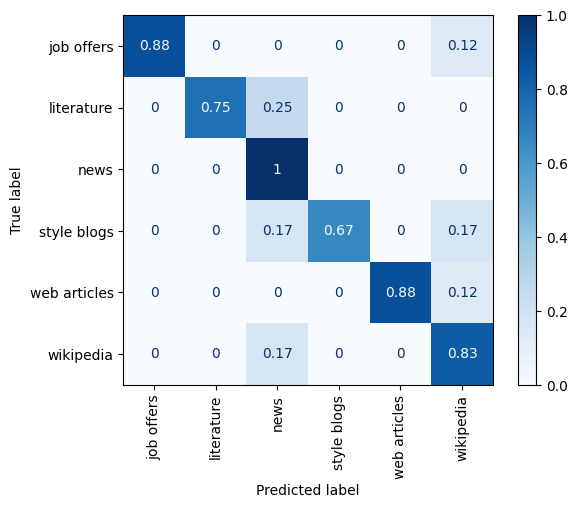}
    \caption{Confusion matrix for text classification task on six categories from the Speakleash library}
    \label{fig:spich2}
\end{figure}

This example also illustrates applications where StyloMetrix vectors outperform large models like BERT-based ones, as the limited sample size per class would not suffice for their fine-tuning.
Another clear advantage of StyloMetrix vectors is their independence from text length. Since the metrics are normalized to the number of tokens in the document, always yielding values between 0 and 1, the issue of overfitting due to varying document lengths between classes is not directly encountered.

\begin{figure}[!ht]
    \centering
    \includegraphics[width=\columnwidth]{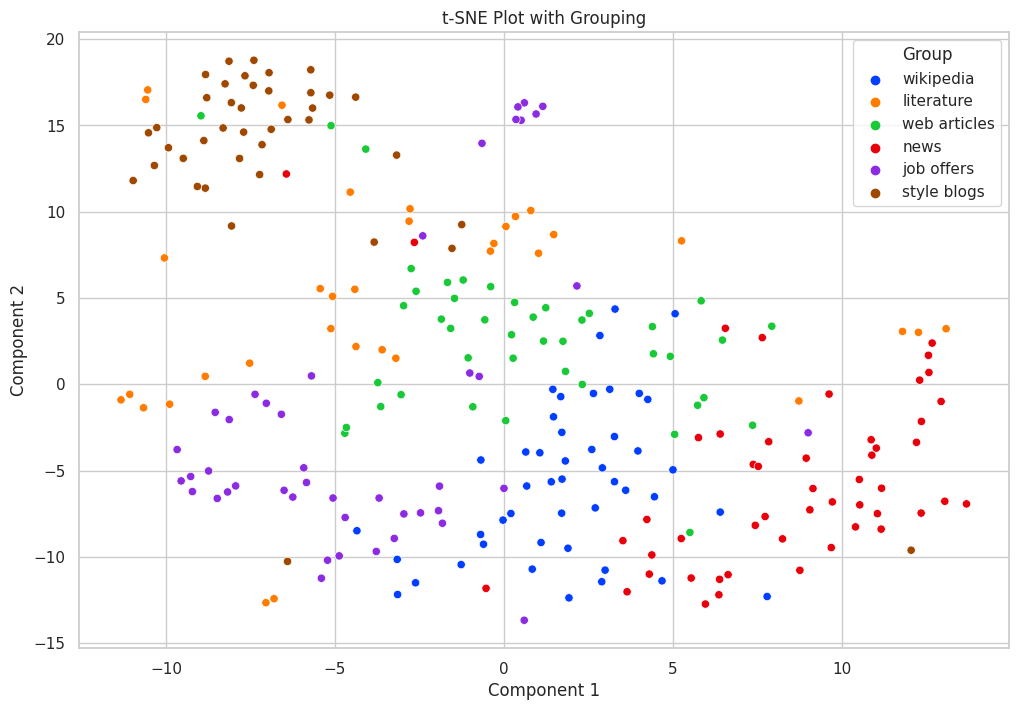}
    \caption{The same six categories text genres from Speakleash library clustered with t-SNE using StyloMetrix vectors}
    \label{fig:spich}
\end{figure}

\subsection{Combination with Deep Learning Embeddings}
One of the most exhaustive analysis has been conducted utilizing the English version of the tool. Our primary concern is to track whether the additional information from the StyloMetrix can contribute to the semantic embeddings produced by Transformers, such as RoBERTa~\cite{zhu-etal-2020-identifying} or BERT, to classify offensive or hate speech.The Null hypothesis claims that the StyloMetrix vectors can enhance the semantic embeddings and provide a more rigorous classification of abusive content. 

With this in mind, at the preliminary stage, we developed the experiments that leverage standard machine learning algorithms such as Random Forest and Logistic Regression. Further, switching to the Transformer models such as HateBERT\footnote{\url{https://huggingface.co/GroNLP/hateBERT}} and RoBERTa. The full list of hyper-parameters and models' layers can be found in Appendix A.

As for the datasets, we have evaluated two most widespread corpora for the hate speech detection: ETHOS\footnote{\url{https://www.bing.com/search?pglt=673&q=ethos+dataste&cvid=c25ed7cede15416a8b9da2708493c2da&aqs=edge..69i57j0j69i64j69i11004.2564j0j1&FORM=ANNAB1&PC=EE23}} and Reddit ~\cite{Gab} datasets. 

The Table~\ref{tab:comp} presents the general tendency of the StyloMetrix vector to enhance the performance of any classification model regardless the dataset. The only exception is the RoBERTa model on the Reddit corpus, where the StyloMetrix decreased the F1 score on hate detection, nevertheless improving the weighted F1 average for both classes. 

\newblock
\begin{table}[!ht]
\centering
\scalebox{1.0}{%
\begin{tabular}{|lll|}
\hline
\multicolumn{1}{|c|}{\textbf{Model}}  
& \multicolumn{1}{c|}{\textbf{weigh. F1}} & \multicolumn{1}{c|}{\textbf{avg. F1}} \\ \hline
\multicolumn{3}{|c|}{\textbf{ETHOS}}                                                                                                           \\ \hline
\multicolumn{1}{|l|}{RoBERTa only}                 & \multicolumn{1}{l|}{0.73}                 & 0.63                                          \\ \hline
\multicolumn{1}{|l|}{RoBERTa with SM}  & \multicolumn{1}{l|}{0.75}                 & 0.67                                          \\ \hline
\multicolumn{1}{|l|}{HateBERT only}                & \multicolumn{1}{l|}{0.77}                 & 0.70                                          \\ \hline
\multicolumn{1}{|l|}{HateBERT with SM} & \multicolumn{1}{l|}{0.81}                 & 0.74                                          \\ \hline
\multicolumn{3}{|c|}{\textbf{Reddit}}                                                                                                          \\ \hline
\multicolumn{1}{|l|}{RoBERTa only}                 & \multicolumn{1}{l|}{0.73}                 & 0.61                                          \\ \hline
\multicolumn{1}{|l|}{RoBERTa with SM}  & \multicolumn{1}{l|}{0.77}                 & 0.60                                          \\ \hline
\multicolumn{1}{|l|}{HateBERT only}                & \multicolumn{1}{l|}{0.79}                 & 0.61                                          \\ \hline
\multicolumn{1}{|l|}{HateBERT with SM} & \multicolumn{1}{l|}{0.80}                 & 0.63                                          \\ \hline
\end{tabular}%
}
\caption{Comparative evaluation of the transformer models with and without the StyloMetrix vectors.}
\label{tab:comp}
\end{table}

\begin{figure}[!ht]
    \centering
    \includegraphics[width=\columnwidth]{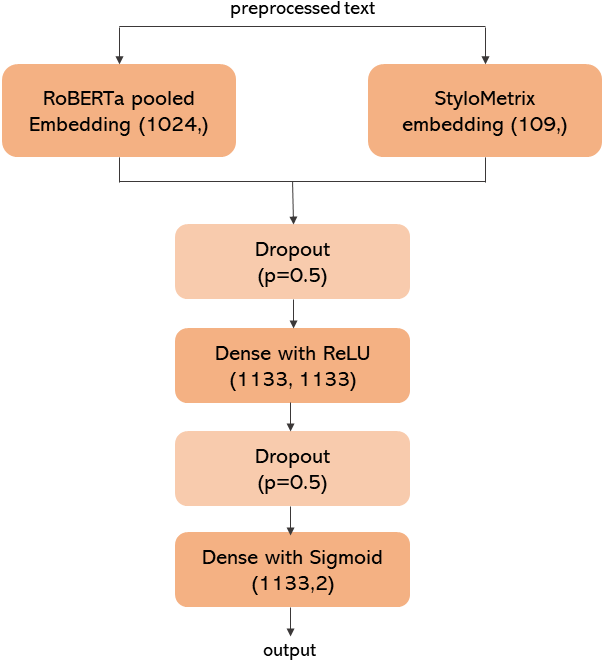}
    \caption{RoBERTa and HateBERT models' layers with StyloMetrix vectors}
    \label{fig:mod}
\end{figure}

An analogous experiment regarding cyberbullying detection was also conducted in the Polish language. The dataset included posts and comments gathered from the Wykop.pl platform, often referred to as the Polish equivalent of Reddit. A detailed description of the dataset is available here \cite{okulska2023ban}. The following experiments were conducted using a transformer model and StyloMetrix vectors:

\begin{enumerate}
    \item The output probabilities from the Polish RoBERTa large model was used as an input to the Logistic Regression (LR) model.
    \item Then fine-tuned RoBERTa large with an additional classification layer was considered. The model used the average of the last hidden state from RoBERTa as a pooled output from the model. This output was activated with ReLU and went to a fully connected layer. The used loss function was BCEWITHLOGITSLOSS (combining Sigmoid layer and the BCELoss function). The learning rate used for the experiment was 1.5e-6. 
    \item Finally, a modification of the fine-tuned RoBERTa with the exact same settings except for the embedding size was considered. Now the StyloMetrix vector has been concatenated with pooled RoBERTa output yielding the inputs of 1133 length (instead of 1024) for the last 2 layers as shown in Fig~\ref{fig:mod}. Additionally, a set of new metrics has been developed dedicated for offensive comments, addressing different types of apostrophe (the figure of speech). 
\end{enumerate}

As shown in Table~\ref{tabres} concatenating the RoBERTa embeddings with the StyloMetrix vectors increases the overall metrics of the model also in Polish language.

This concise description of experiments presents one of many possibilities to employ the StyloMetrix vectors with machine learning and deep learning algorithms, enhancing their operation and yielding stronger predictions.

\begin{table}

\centering
\begin{tabular}{|l|l|l|l|l|}
\hline
\textbf{Model} & \textbf{Rec} & \textbf{F1}  \\
\hline \hline
Pre-trained RoBERTa                   & 0.88   & 0.90    \\
RoBERTa fine-tuned                               & 0.93   & 0.93     \\
RoBERTa with SM       & \textbf{0.94}   & \textbf{0.94}     \\
\hline
\end{tabular}
\caption{Results of the models for the Wykop content moderation task -- Recall and weighted F1 score}
\label{tabres}
\end{table}

\section{Previous Applications of StyloMetrix}

StyloMetrix was initially applied in NLP tasks to classify text genres with and without adult content and to analyze sentiment in customer reviews~\cite{okulska2022}. Experiments using StyloMetrix vectors as input for Random Forests were conducted on small datasets. In the case of genre classification, the results outperformed classification results obtained with deep learning models on the same data, achieving an mean accuracy of 93.5\% and balanced results across classes.

Text representations generated with StyloMetrix were used alongside HerBERT-based embeddings to detect genres and persuasion techniques in Polish~\cite{modzelewski-etal-2023-dshacker}. The experiments employed classical machine learning models: LightGBM, XGBoost, and logistic regression. The approach adopted for detecting persuasion techniques resulted in a third-place finish in SemEval-2023 Task 3~\cite{piskorski-etal-2023-semeval}. 

In the realm of computational linguistics, StyloMetrix text analysis was used in a profound linguistic analysis of offensive language and hate speech from Wykop.pl web service~\cite{okulskakolos2024}. The tool was also used for the task of author classification in Polish poetry~\cite{okulskakolosskibski2024}. These applications underscore the tool's utility in addressing complex societal issues and fostering a deeper understanding of online language dynamics, as well adapting to diverse text analysis challenges for literary research purposes. The efficiency of StyloMetrix for the Ukrainian language was analyzed in detail by \citet{stetsenko2023grammar}. The authors provide a more in-depth analysis of the language characteristics, some limitations of the tool and offer an overview on the StyloMetrix usage as an explainable model for text classification.




\section{Limitations}
The reasons for divergences in the tool output may be several. SpaCy models have limited accuracy, and the errors in model performance may propagate and skew metrics values (although some mistakes are corrected as an element of the StyloMetrix pipe). Language models do not perform well on contaminated data -- typos, imperfect data sources (e.g., OCR), colloquial language or inconsistent grammar may result in incorrect labeling and disrupt dependency parsing. Since the metrics calculations rely on substantive knowledge, they have to be implemented into the system by an expert in the field, which opens the possibility for human error. Therefore in sensitive (e.g., jurisdictional) applications, it is recommended to use the function of debugging information output for manual check. In the phase of data analysis, vectors of short texts may turn out to be too scarce in non-empty values to represent a writing style, and therefore the classification may be ineffective. Low accuracy could also occur when different texts having similar structure. There will be potentially classes where solely stylometric representation will still require support from a semantic classifier to yield fully satisfying results.

\section{Conclusion}

There are many different methods for vectorizing textual data, and it is crucial to select the appropriate approach for a given task. Stylometric approach shows a great potential for solving various NLP problems. The open-source multilingual StyloMetrix vectors presented in the paper offer:
\begin{enumerate}
    \item Normalized advanced stylometric statistics with a fixed length regardless of the input text size.
    \item The ability to serve as input for text classification tasks, such as content, genre, authorship, or style classification. They are also effective in semantic classification, for instance, in detecting hate speech.
    \item The potential to enhance existing text embeddings, leading to higher classification effectiveness.
    \item Model's explainability, as they perform well on simple tree-based models like Random Forest, with each value in the StyloMetrix vector (each model's feature) directly explaining a specific grammatical or lexical pattern.
    \item An open-source for advanced linguistic analysis of the entire text corpora.
    \item Availability for up to four languages: English, Polish, Ukrainian, and Russian.
\end{enumerate}

\bibliography{custom}
\bibliographystyle{acl_natbib}

\appendix

\section{Model hyperparameters}
For fine-tuning the transformers: Adam(learning rate=5e-5), Epochs: 3; Batch size: 16; Class weights added;

For Random Forest: number of estimators = 200, max\_features = 'auto', max\_depth = 183, n\_jobs=-1, random\_state=22

\section{Code examples}
Code excerpts for an exemplary metric in the Syntactic Forms category for the Polish language: 
\begin{lstlisting}[language=Python]
class SY_S_NEG(Metric):
    category = Syntaktyka
    name_en = 'negative_sentences'
    name_local ='zdania przeczace'

    def count(doc):
        neg_sents = []
        for sent in doc.sents:
            if any(token.dep_ 
            in ["ROOT", "ccomp", 
            "cop"] and token.pos_ 
            in ["VERB", "AUX"] for 
            token in sent) and 
            any(token.dep_ == 
            "advmod:neg" and 
            token.pos_ == "PART" 
            for token in sent):
                neg_sents.append(sent)
        debug = [token for sent in 
        neg_sents for token in sent]
        result = len(debug)
        return ratio(result, len(doc)) 
\end{lstlisting}

\end{document}